\documentclass[11pt]{article}
\usepackage{coling2016}
\usepackage{times}
\usepackage{url}
\usepackage{latexsym}

\usepackage{amsmath,bm}
\usepackage{amsfonts}
\usepackage{mathrsfs}
\usepackage{amsbsy,amsthm,amssymb}

\usepackage{graphicx}
\usepackage{xcolor}
\usepackage{multirow}
\usepackage{paralist}
\usepackage{subfigure}
\usepackage{arydshln}
\usepackage{CJK}
\usepackage{fancyhdr}

\newcommand{\mysmall}{\fontsize{10pt}{10pt}}

\title{Sequence to Backward and Forward Sequences: A Content-Introducing Approach to Generative Short-Text Conversation}

\author{Lili Mou,$^1$ Yiping Song,$^2$ Rui Yan,$^3$ Ge Li,$^{1,*}$ Lu Zhang,$^1$ Zhi Jin$^{1,*}$\\
\normalsize$^1$Key Laboratory of High Confidence Software Technologies (Peking University), MoE, China\\
\normalsize Institute of Software, Peking University, China\quad $^*$Corresponding authors\\
\normalsize $^2$Institute of Network Computing and Information Systems, Peking Univerity, China\\
\normalsize $^3$Institute of Computer Science and Technology, Peking University, China\\
\mysmall \texttt{doublepower.mou@gmail.com}\\[-.1cm]
\mysmall \texttt{\{songyiping,ruiyan,lige,zhanglucs,zhijin\}@pku.edu.cn}\\
}

\date{}

\newcommand{\seqseq}{\texttt{seq2seq}}

\newcommand{\seqbf}{\texttt{seq2BF}}
\newcommand{\pmi}{\operatorname{PMI}}
\newcommand{\argmax}{\operatorname*{argmax}}

\begin{document}
\begin{CJK*}{UTF8}{gkai}

\maketitle
\begin{abstract}
Using neural networks to generate replies in human-computer dialogue systems is attracting increasing attention over the past few years. However, the performance is not satisfactory: the neural network tends to generate safe, universally relevant replies which carry little meaning. In this paper, we propose a content-introducing approach to neural network-based generative dialogue systems. We first use pointwise mutual information (PMI) to predict a noun as a keyword, reflecting the main gist of the reply. We then propose \seqbf, a ``\textit{sequence to backward and forward sequences}'' model, which generates a reply containing the given keyword. Experimental results show that our approach significantly outperforms traditional sequence-to-sequence models in terms of human evaluation and the entropy measure, and that the predicted keyword can appear at an appropriate position in the reply.
\end{abstract}

\section{Introduction}
\blfootnote{
    %
    %
    %
    %
    %
    %
    \hspace{-0.65cm}  
    This work is licenced under a Creative Commons
    Attribution 4.0 International License.
    License details:
    \url{http://creativecommons.org/licenses/by/4.0/}
}
Automatic human-computer conversation is a hot research topic in natural language processing (NLP). In past decades, researchers have developed various rule- or template-based systems, which are typically in vertical domains, e.g., transportation \cite{train} and education \cite{tutor}. In the open domain, data-driven approaches play an important role, because the diversity and uncertainty make it virtually impossible for humans to design rules or templates. \newcite{retrieval1} and \newcite{retrieval2} use information retrieval methods to search for a reply from a pre-constructed database; \newcite{SMTdialog} formalize conversation as a statistical machine translation task.

Recently, the renewed prosperity of neural networks brings new opportunities to open-domain conversation \cite{NNdialog1,NRM,HierarchicalDialog,diversity}. In these studies, researchers leverage sequence-to-sequence (\seqseq) models to encode a \textit{query} (user-issued utterance) as a vector and to decode the vector into a \textit{reply}. In both encoders and decoders, an RNN keeps one or a few hidden layers; at each time step, it reads a word and changes
its state accordingly. RNNs are believed to be well capable of modeling word sequences, benefiting machine translation \cite{seq2seq}, abstractive summarization \cite{seq2seqSummarization} and other tasks of natural language generation. Contrary to retrieval methods, neural network-based conversation systems are \textit{generative} in that they can synthesize new utterances; results in the literature also show the superiority of \seqseq\ to phrase-based machine translation for dialogue systems \cite{NRM}. In our study, we focus on neural network-based generative \textit{short-text conversation}, where we do not consider context information, following \newcite{retrieval2} and \newcite{NRM}.

Despite these, neural networks' performance is far from satisfactory in human-computer conversation. A notorious problem is the \textit{universal reply}: the RNN prefers to generate  safe, universally relevant sentences with little meaning, e.g., ``something'' \cite{HierarchicalDialog} and ``I don't know'' \cite{diversity}. One problem may lie in the objective of decoding. If we choose a reply with the maximal estimated probability (either greedily or with beam search), it is probable to obtain such universal replies, because they do appear frequently in the training set. Another potential problem is that, the query may not convey sufficient information for the reply, and thus the encoder in \seqseq\ is less likely to obtain an informative enough vector for decoding.

In this paper, we propose a content-introducing approach to generative short-text conversation systems, where a reply is generated in a two-step fashion: (1) First, we predict a keyword, that is, a noun reflecting the main gist of the reply. This step does not capture complicated semantic and syntactic aspects of natural language, but estimates a keyword with the highest pointwise mutual information against query words. The keyword candidates are further restricted to nouns, which are not as probable as universal words (e.g., \textit{I} and \textit{you}), but can introduce substantial content to reply generation. (2) We then use a modified encoder-decoder model to synthesize a sentence containing the keyword. In traditional \seqseq, the decoder generates the reply from the first word to the last in sequence, which prevents introducing certain content (i.e., a given word) to the reply. To tackle this problem, we propose \seqbf, a novel ``\textit{sequence to backward and forward sequences}'' model, based on our previous work of backward and forward language modeling \cite{BF}. The \seqbf\ model
decodes a reply starting from a given word, and generates the remaining previous and future words subsequently. In this way, the predicted keyword can appear at an arbitrary position in the generated reply.

The rest of this paper is organized as follows. Section~\ref{sec:approach} describes the proposed approach; Section~\ref{sec:exp} presents experimental results. Section~\ref{sec:related} briefly reviews related work in the literature. Finally we conclude our paper and discuss future work in Section~\ref{sec:conclusion}.

\section{Our Approach}\label{sec:approach}
In this section, we present our content-introducing generative dialogue system in detail. Subsection~\ref{ss:overview} provides an overview of our approach, Subsection~\ref{ss:keyword} introduces the keyword predictor, and Subsection~\ref{ss:seqbf} elaborates the proposed \seqbf\ model. We describe training methods in Subsection~\ref{ss:train}.

\subsection{Overview}\label{ss:overview}

\begin{figure*}[!t]
\centering
\includegraphics[width=.85\textwidth]{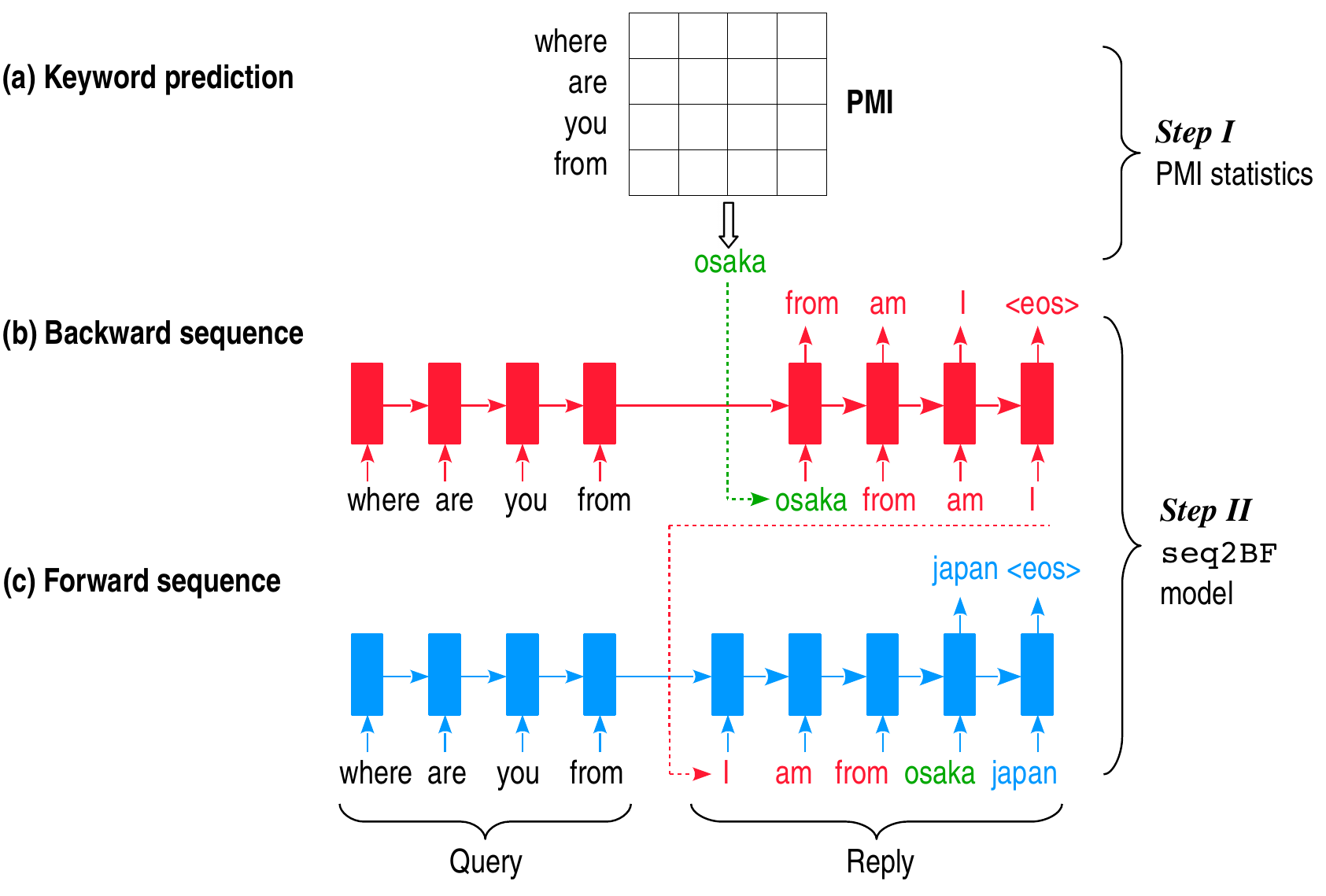}
\caption{An overview of our content-introducing approach to generative dialogue systems.}\label{fig:overview}
\end{figure*}

Figure~\ref{fig:overview} depicts the overall architecture of our approach, which comprises two main steps:
\begin{itemize}
\item[\textbf{\emph{Step I}}:] We first use PMI to predict a keyword for the reply, as shown in
 Figure~\ref{fig:overview}a.
\item[\textbf{\emph{Step II}}:] After keyword prediction, we generate a reply conditioned on the keyword as well as the query. More specifically, we propose the \seqbf\ model, which generates the backward half of the sequence (Figure~\ref{fig:overview}b) and then the forward half (Figure~\ref{fig:overview}c).
\end{itemize}

Notice that,  the RNNs in Step II do not share parameters (indicated by different colors in the figure) because they differ significantly from each other. Moreover, the encoder and decoder do not share parameters either, which is standard in \seqseq. For clarity, we do not assign different colors for encoders and decoders, but separate them with a long arrow in Figures~\ref{fig:overview}b and~\ref{fig:overview}c.

\subsection{Keyword Prediction}\label{ss:keyword}
In this step, we use pointwise mutual information (PMI) to predict a keyword for the reply. We leverage such surface statistics because this step outputs a single keyword, which does not capture complicated syntax and semantics of queries and replies. Our goal of content introducing is to suggest a word that is especially suited to the query, instead of predicting a most likely (common) word. Hence, the pointwise mutual information is an appropriate statistic for keyword prediction.

Formally, we compute PMI of a query word $w_q$ and a reply word $w_r$ using a large training corpus by
\begin{equation}
\pmi(w_q, w_r)=\log\dfrac{p(w_q,w_r)}{p(w_q)p(w_r)}=\log \dfrac{p(w_q|w_r)}{p(w_q)}
\end{equation}

When predicting, we choose the keyword $w_r^*$ with the highest PMI score against query words $w_{q_1}, \cdots w_{q_n}$, i.e., $w_r^*=\argmax_{w_r} \pmi(w_{q_1}\cdots w_{q_n}, w_r)$, where
\begin{align}
&\pmi(w_{q_1}\cdots w_{q_n}, w_r)=\log \dfrac{p(w_{q_1}\cdots w_{q_n}|w_r)}{p(w_{q_1}\cdots w_{q_n})}\\
\approx&\log \dfrac{\prod_{i=1}^n p(w_{q_i}|w_r)}{\prod_{i=1}^n p(w_{q_i})}=\sum_{i=1}^n \log\dfrac{p(w_{q_i}|w_r)}{p(w_{q_i})}=\sum_{i=1}^n \pmi(w_{q_i},w_r)
\end{align}
The approximation is due to the independency assumptions of both the prior distribution $p(w_{q_i})$ and posterior distribution $p(w_{q_i}|w_r)$. While the two assumptions may not be true, we use them in a pragmatic way so that the word-level PMI is additive for a whole utterance. Experiments show that this treatment generally works well.

Different from choosing the most likely word, PMI penalizes a common word by dividing its prior probability; hence, PMI prefers a word that is most ``mutually informative'' with the query. Moreover, we manually restrict keyword candidates to nouns, so that this step can introduce substantial content to reply generation, which will be discussed in the next part.

\subsection{The \seqbf\ Model}\label{ss:seqbf}

To insert the predicted keyword into sequence generation, we cannot use the traditional \seqseq\ model. In existing approaches, we usually decompose the probability of an output sentence $\bm r=r_1 r_2 \cdots r_m$ given an input sentence $\bm q=q_1q_2\cdots q_n$ by
\begin{align}
p(r_1,\cdots, r_m|\bm q)=p(r_1|\bm q)p(r_2|r_1,\bm q)\cdots p(r_m|r_1\cdots r_{m-1},\bm q)=\prod_{i=1}^m p(r_i|r_1\cdots r_{i-1},\bm q)
\end{align}
The output sentence is thus predicted in sequence from $r_1$ up to $r_m$ either greedily or with beam search. I personally believe such decomposition is mainly inspired by the observation that humans always say a sentence from the beginning to the end.

However, in our content-introducing approach to generative dialogue systems, the predicted keyword could appear at the beginning ($r_1$), the middle ($r_2$ to $r_{m-1}$), or the end ($r_m$) of the reply. It is then natural to decompose the probability starting from the given word. In particular, the predicted  keyword $k$ splits a reply into two (sub-)sequences:
\begin{eqnarray}\nonumber
\text{Backward sequence:}&&  r_{r-1}, \cdots, r_1\\ \nonumber
\text{Forward sequence:}&& r_{k+1}, \cdots, r_m
\end{eqnarray}
and the joint probability of remaining words can be written as
\begin{align}
&p\!\left(\!\begin{array}{l}
r_{k-1}\cdots r_1\\\hdashline
r_{k+1}\cdots r_m
\end{array}\Bigg|r_k,\bm q\right)=\prod_{i=1}^{k-1}p^{\text{(bw)}}(r_{k-i}|r_k,\bm q,\cdot)
\prod_{i=1}^{m-k}p^{\text{(fw)}}(r_{k+i}|r_k,\bm q,\cdot)\label{eqn:BF}
\end{align}
where $p(\text{-\ -\ -}|r_k,\bm q)$ refers to the probability of the backward and forward subsequences given the split word $r_k$ and an encoded query $\bm q$.
Notice that both the backward and forward sequence generators include a wildcard allowing rich inner-subsequence and/or inter-subsequence dependencies. In our previous study of backward-and-forward~(B/F) language modeling~\cite{BF}, we propose three variants: (1) \textbf{sep-B/F}: The backward and forward sequences are generated separately. (2) \textbf{syn-B/F}: The backward and forward sequences are generated synchronously using a single RNN, two output layers at each time step for the two sequences. (3) \textbf{asyn-B/F}: The two sequences are generated asynchronously, that is, we first generate the backward ``half'' sequence, conditioned on which we then generate the forward ``half.'' Our previous experiments show the asyn-B/F is the most natural way of modeling backward and forward sequences, and thus we adopt this variant in \seqbf.

Specifically, our \seqbf\ model works as follows. A \seqseq\ neural network encodes a query and decodes a ``half'' reply, that is, the first set of factors in Equation~\ref{eqn:BF} becomes $p^{\text{(bw)}}(r_{k-i}|r_k,\bm q,\cdot)=p^{\text{(bw)}}(r_{k-i}|r_k\cdots r_{k-i+1},\bm q)$, where $1\le i\le k-1$. The decoder here outputs words in a reversed order from $r_{k-1}$, $r_{k-2}$ to $r_1$, so that the reversed ``half'' sequence is fluent with respect to the given word, at least from a mathematical perspective. (Please see Figure~\ref{fig:overview}b.)

Then another \seqseq\ model encodes the query again, but decodes the entire reply, provided that the first half of the reply is given (Figure~\ref{fig:overview}c), i.e., $p^{\text{(fw)}}(r_{k+i}|r_k,\bm q,\cdot)=p^{\text{(fw)}}(r_{k+i}|r_1\cdots r_k\cdots r_{k+i-1},\bm q)$, $(1\le i\le m-k)$. Here, the forward generator is aware of the backward half sequence $r_1\cdots r_{k-1}$, where the word order is reversed again, so that they are in a normal order for fluent forward generation.

 In both backward and forward \seqseq\ models, we use RNNs with gated recurrent units (GRUs) for information processing~\cite{GRU}, given by
\begin{align}
\bm r_t&=\sigma(W_r \bm w_t + U_r\bm h_{t-1}+\bm b_r)\\
\bm z_t&=\sigma(W_z \bm w_t+U_r\bm h_{t-1}+\bm b_z)\\
\widetilde{\bm h}_t&=\tanh\Big(W_h\bm w_t+U_h(\bm r\circ\bm h_{t-1})+\bm b_h\Big)\\
\bm h_t &= (1-\bm z_t)\circ \bm h_{t-1}+\bm z_t\circ\widetilde{\bm h}_t
\end{align}
where $W$'s and $U$'s are weights and $\bm b$'s are bias terms. $\bm w_t$ is the word embedding; $\bm h_t$ is the hidden state at the time step $t$. ``$\circ$'' denotes element-wise product.

\subsection{Model Training}\label{ss:train}

Training (i.e., parameter estimation) is always a most important thing in the neural network regime, and oftentimes, problems arise when we prepare the dataset.

Fortunately, the \seqbf\ model can be trained without additional labels. We randomly sample a word in a reply as the split word, take the first half, and reverse its word order; in this way, we obtain the training data for the backward sequence generator. The forward sequence generator is essentially a \seqseq\ encoder and decoder from queries to replies. The difference between the pure \seqseq\ and the forward generator of \seqbf\ lies in the inference stage: in our scenario, we ignore the query-reply \seqseq\ generator's output at the beginning steps, but feed it with the ``half'' reply obtained by our backward sequence generator as well as the predicted keyword (red and green words in Figure~\ref{fig:overview}c); then we let the \seqseq\ model generate remaining future words (blue words in Figure~\ref{fig:overview}c).

It should be emphasized that the backward sequence generator requires ``half'' replies starting from the split word as training data, and that we cannot train the model with a full reversed sentence. Otherwise, the backward part will undesirably generate an entire reversed reply, and the forward part cannot add much to it.

\section{Experiments}\label{sec:exp}

\subsection{Dataset}
We evaluated our approach on a Chinese dataset of human conversation crawled from the Baidu Tieba\footnote{http://tieba.baidu.com} forum. We used 500,000 query-reply pairs to train the \seqbf\ model. We had another unseen 2000 and 27,871 samples for validation and testing, respectively. To obtain PMI statistics in the first step (Figure~\ref{fig:overview}a) of our method, we use a much larger dataset containing 100M query-reply pairs.

Chinese language is different from English in that a Chinese character carries more semantics than an alphabet. For example, the characters {黑} and {板} mean ``black'' and ``board'' in English, respectively; the term {黑板} means ``blackboard.'' Because we have far more Chinese terms than English words, our \seqbf\ is trained in the character level out of efficiency concerns. But we train the keyword predictor with noun phrases (Chinese terms), by noticing that \textit{blackboard} is different from \textit{board}, despite some subtle relations. Fortunately, the two granularities can be integrated together straightforwardly: during backward sequence generation, we only need to condition the model on the character sequence in the key term instead of a single keyword, that is, we have several green inputs in Figure~\ref{fig:overview}b. In our study, we kept 2.5k noun terms as candidate keywords and 4k characters for \seqbf\ generation.

\subsection{Hyperparameters}

In our experiments, word embeddings and recurrent layers were 500d. We used rmsprop to optimize all parameters except embeddings, with initial weights uniformly sampled from $[-.08,.08]$, initial learning rate $0.002$, moving average decay $0.99$, and a
damping term $\epsilon=10^{-8}$.  Because word embeddings are sparse in use \cite{comparative}, we optimized embeddings asynchronously by stochastic gradient descent with the learning rate divided by $\sqrt{\epsilon}$.
We set the mini-batch size to 50. These values were mostly chosen empirically by following \newcite{visualizeRNN} and \newcite{BF}; they generally work well in our scenarios. We did not tune the hyperparameters in this paper, but are willing to explore their roles in dialogue generation as future work.

The validation set (containing 2k query-reply samples) was used for early stop only. We chose the parameters yielding the highest character-level BLEU-2 score on our validation set.

\subsection{Performance}

We evaluated our results in terms of the following criteria:
\begin{itemize}
\item \textbf{Human Evaluation}. We had six volunteers\footnote{All volunteers are well-educated native speakers of Chinese and have received a Bachelor's degree or above.} to annotate the results of our content-introducing \seqbf\ and baselines. To ensure the quality of human evaluation, we randomly sampled 200 queries and replies in the test set. The samples and volunteers were further split into two equal-sized groups of different annotation protocols:
\begin{compactitem}
\item \textit{Pointwise annotation}. The volunteers were asked to annotate a score indicating the appropriateness of a reply to a given query: 0 = bad reply, 2 = good reply, and 1 = borderline.
\item \textit{Pairwise annotation}. Given a certain query, the volunteers were asked to judge whether  pure \seqseq\ is better than, equal to, or worse than \seqbf\ (with content introducing). If they could not understand both replies, they were asked to choose ``equal.''
\end{compactitem}

All our human evaluation were conducted in a random, blind fashion, i.e., we randomly shuffled the samples and volunteers did not know which system generated a particular reply.
Notice that we did not define what a ``good'' or ``bad'' reply is; otherwise, the annotation may be biased towards certain systems. Rather, annotators had their own subjective discretion. While criteria may differ from person to person, the ranking of average scores reflects the comparison of different dialog systems. (In our study, the ranking is consistent among all six annotators.)

\item \textbf{Length}. The length of an utterance is an objective, surface metric that reflects the substance of a generated reply.
\item \textbf{Entropy}. Entropy is another objective metric, which shows the serendipity of a reply by measuring the amount of information contained in the utterance.  We computed the average character-level entropy, given by
\begin{equation}
-\dfrac 1{|R|}\sum_{w\in R}\log_2 p(w)
\end{equation}
where $R$ refers to all replies, $|R|$ is the number of words in all replies, and $p(\cdot)$ is the unigram probability of a character in the training set.
\end{itemize}

The latter two metrics are ``intrinsic,'' by which we mean no reference (groundtruth reply) is needed to compute the metric. They are used in \newcite{variationalDialog}.\footnote{The entropy equation in \newcite[v3]{variationalDialog}  has a minor error. We confirmed this with Serban via personal emails.} In our experiments, objective metrics were assessed with all test samples.

\begin{table}[!t]
\centering
\begin{minipage}{.52\textwidth}
\begin{tabular}{|l|r|rr|}
\multicolumn{4}{c}{(a)}\\
\hline
\textbf{Method}&\textbf{PointHuman} & \textbf{Length} & \textbf{Entropy}\\
\hline\hline
\seqseq  & 0.58 & \textbf{5.61} & 6.960\\
\seqbf$_-$ & 0.46 & 5.60 & 6.971\\
\seqbf$_+$ & \textbf{0.67} & {5.31} & \textbf{9.139} \\
\hline
Groundtruth &   --  & 9.19 & 8.832\\
\hline
\end{tabular}
\end{minipage}~~~~~~
\begin{minipage}{.35\textwidth}
\begin{tabular}{|l|ccc|}
\multicolumn{4}{c}{(b)}\\
\hline
\multicolumn{4}{|c|}{\textbf{PairHuman}} \\
\hline
\textbf{Method}     & \textbf{Wins} & \textbf{Ties} & \textbf{Loses}\\
\hline
\hline
\seqseq    & 24.7   &  26.0  & 49.3 \\
\seqbf$_+$      & \textbf{49.3}   &  26.0  & 24.7\\
\hline
\end{tabular}
\end{minipage}
\caption{The performance of our content-introducing \seqbf\  (denoted as \seqbf$_+$) dialogue system in comparison with pure \seqseq\ and \seqbf\ without predicted keywords (\seqbf$_-$). \textbf{PointHuman}: Pointwise human evaluation. (Annotator agreement: std=0.33, Fleiss' $\kappa$=0.27.) \textbf{PairHuman}: Pairwise human evaluation, which shows the percentage at which a system wins, loses, or ties in comparison with the other ($\kappa$=0.29).}
\label{tab:performance}
\end{table}
We do not include BLEU scores as our evaluation criteria, which are used in \newcite{diversity}. As a pilot study, we also asked two volunteers to write their own replies to 50+ queries. One obtained 1.69 BLEU-4 score. (The result is lower than 1.74 obtained by an automatic dialogue system in \newcite{diversity}; this may be caused by different datasets and languages.) What surprises us is that the other volunteer obtained 0 BLEU-2 score, indicating that no bi-gram overlaps between his replies and the references. This result provides evidence of the diversity among human replies, and thus we abandoned BLEU scores as evaluation criteria. We reported this case study in our paper so as to shed more light on the research of evaluation metrics in dialogue systems.

Table~\ref{tab:performance} compares our content-introducing \seqbf\ with the \seqseq\ baseline. Also included is the \seqbf\ model without content introducing, that is, we slacked the keyword constraint in the backward sequence generator (Figure~\ref{fig:overview}b), and let it generate a reversed half reply freely.

As seen in Table~\ref{tab:performance}a, the content-introducing \seqbf\ model yields the highest pointwise human annotation result. For \seqbf\ without content introducing, we obtain low scores, showing that artificially splitting a sequence into two parts itself is not a fancy way of modeling natural language sentences. However, given a keyword predicted by PMI statistics, the backward and forward sequence generation can significantly improve the dialogue system in comparison with pure \seqseq. In the pairwise annotation protocol, the content-introducing \seqbf\ model also outperforms traditional \seqseq\ to a large extent. We computed standard deviation (0.33) and discrete-class Fleiss' $\kappa$ scores (0.27--0.29), showing fair annotator agreement \cite{kappa}. While the agreement is comparable to previous results, e.g., 0.2--0.4 reported in \newcite{NRM}, we find the disagreement mainly comes from the discretion of the ``goodness'' of a reply (so that the annotation complies with the subjective nature of human evaluation). All annotators yielded the same system-level ranking order, providing consistent evidence of the effectiveness of our approach.

Regarding intrinsic metrics, our \seqbf\ with a predicted keyword generates slightly shorter replies than \seqseq, but contains far richer information, as the entropy increases by 30\%. The results verify that content introducing is particularly useful in generative human-computer dialogue systems.

\subsection{Case Studies and Discussion}
\begin{table*}[!t]
\centering
\begin{tabular}{|r|ll|}
\hline
               & \textbf{Chinese} & \textbf{English (translated)}\\
\hline
Query & 李有男友公开过了  & It's known that \textit{Li}$^\dag$ has a boyfriend.\\
Grountruth & 都已经分了\ 之前李的贴吧都在讨论了\!\!\!\!\!      & Broken up. There's discussion in her \textit{Tieba}.$^\ddag$ \\
\cline{2-3}
\seqseq         & 我是男的 & I am a male\\
\seqbf$_-$      & 我的男友 & My boyfried\\
\seqbf$_+$      & 有\textbf{绯闻}男友 & Has a \textbf{rumor}ed boyfried\\
\hline
Query          & 人大复试飘过  & Passed second-round exam of Renming Univ.\\
Grountruth     & 这么牛，什么专业    & Cool, what's your major\\
\cline{2-3}
\seqseq         & 我也是！ & Me too!\\
\seqbf$_-$      & 我也是！ & Me too!\\
\seqbf$_+$      & \textbf{分数}是什么& What is your \textbf{score}\\
\hline
Query          & 挺漂亮的\ 祝福祝福 & So beautiful, congratulations\\
Grountruth     &  下手慢了哈。现在    & You're late till now\\
\cline{2-3}
\seqseq         & 谢谢 & Thank you \\
\seqbf$_-$      & 是的，谢谢 & Yes, thank you\\
\seqbf$_+$      & 谢谢\textbf{夸奖} &  Thank you for \textbf{praising}\\
\hline
Query          & 我要换头像了！ & Want to change a photo\\
Grountruth   &  打算换成什么啊～   & What do you like...\\
\cline{2-3}
\seqseq         &  我是你的头像 & I'm in your photo\\
\seqbf$_-$      &  我是你的头像  & I'm in your photo\\
\seqbf$_+$      & 第一张\textbf{图像}是谁& Who is in your first \textbf{photo}\\
\hline

\end{tabular}
\caption{Examples of generated replies. Predicted keywords are in \textbf{bold}. $^\dag$\textit{Li} is a Chinese actress; we anonymized her firstname. $^\ddag$\textit{Tieba}  is a Chinese forum where our datasets are crawled.}
\label{tab:example}
\end{table*}
We provide case studies in Table~\ref{tab:example}.\footnote{Due to Chinese-English translation, some characteristics cannot be fully presented in the English text, e.g., the position of the given word, the length of the reply, and even the part-of-speech of a word. Nevertheless, we present the predicted keyword and its translated counterpart in bold and thus the aforementioned characteristics can be visually demonstrated to some extent.} As we see, the \seqseq\ responder prefers safe, universally relevant utterances like ``me too.'' In these examples, the replies generally match the queries in meaning, but such universal replies are too boring and thus undesirable in real applications.
In the \seqbf\ model with content introducing, we predict a keyword of the reply with PMI. This yields meaningful words/terms like \textit{rumor} and \textit{score}, serving as the gist of the reply. Then the \seqbf\ model generates previous and future words to obtain a more complete utterance (maybe not a whole sentence because of the casualness in human conversation). The proposed ``backward and forward sequences'' mechanism ensures that the predicted keyword can appear at an arbitrary position in the utterance.

\begin{table}[!t]
\centering
\begin{tabular}{|l|cc|cc|}
\hline
\multirow{2}{*}{\textbf{Model}}        &  \multirow{2}{*}{\seqseq}  &   \multirow{2}{*}{\seqbf$_-$}   & \multicolumn{2}{c|}{\seqbf$_+$}\\
        &                             &      & \makebox[2cm]{keyword}   & \makebox[2cm]{remaining} \\
\hline
\textbf{Entropy} &  6.960                     &  6.971   & 11.630     & 7.422\\
\hline
\end{tabular}
\caption{Fine-grained analysis of character-level entropy. In the \seqbf$_+$ model, we analyze the average entropy of keywords and remaining words separately.}\label{tab:entropy}
\end{table}
We delve deep into the question: why \seqseq\ models (or variants) tend to generate universal replies? We may have two conjectures: (1) The \seqseq\ model cannot capture rich enough semantics other than ``yes,'' ``me too,'' etc. (2) The sequence generator is able to capture rich semantics, but starting from a high-level universal word at the beginning, it is unlikely to fall into concrete concepts.

We present in Table~\ref{tab:entropy} the average character-level entropy of keywords and non-keywords.
We find that, provided with a noun term, \seqbf\ can generate meaningful remaining words (keyword excluded) with an entropy of 7.422, higher than 6.971 given by \seqbf\ without keywords.
Noticing that the \seqbf\ model is exactly the same in content-introducing and non-content-introducing settings, we believe the second conjecture holds. Choosing the most likely reply yields universal utterances; moreover, RNN sequence generators are reluctant to introduce concrete concepts, provided with one or a few universal words/terms (like \textit{I} and \textit{you}) that are greedily chosen at the beginning.

Fortunately, our content-introducing \seqbf\ works in an opposite fashion. We first predict a meaningful but not that probable noun term as the keyword; then we feed \seqbf\ with such concrete keyword that provides substantial content. In this way, our approach significantly outperforms pure \seqseq\ generation in short-text conversation systems.

\section{Related Work}\label{sec:related}
\subsection{Dialogue Systems}
Automatic human-computer conversation has long attracted the attention of researchers. In early decades, people design rule- or template-based systems, but they are mainly in vertical domains~\cite{train,museum}. Although such approaches can also be extended to the open domain \cite{7patterns}, their generated sentences are subject to 7 predefined forms and thus are highly restricted. For open dialogues, researchers have applied data-driven approaches, including retrieval methods~\cite{retrieval1,retrieval2}, phrase-based machine translation~\cite{SMTdialog}, and recurrent neural networks~\cite{BoWdialog,NRM}.

A hot research topic in human-computer conversation is mixed-initiative systems, for example, the TRAINS-95 system for route planning~\cite{train} and AutoTutor for learner advising~\cite{tutor}. \newcite{stalematebreaker} propose a proactive dialogue system that can introduce new content when a stalemate occurs. The system is chatbot-like and in the open domain; an external knowledge base is used for searching related entities as new content. They propose a random walk-like reranking algorithm based on retrieval results. Different from \newcite{stalematebreaker}'s work, our paper addresses the problem of content introducing in open-domain generative dialogue systems.

\subsection{Neural Networks for Sentence Generation}

\newcite{seq2seq} propose \seqseq\ for machine translation; the idea is to encode a source sentence as a vector by a recurrent neural network (RNN) and to decode the vector to a target sentence by another RNN. \newcite{attention} enhance it with an attention mechanism. These approaches largely benefit natural language generation tasks such as abstractive summarization \cite{seq2seqSummarization}, question answering \cite{genQA}, and poetry generation~\cite{poem}.

For neural network-based dialogue systems, \newcite{BoWdialog} summarize a query and context as bag-of-words features, based on which an RNN decodes the reply. \newcite{NRM} generate replies for short-text conversation by \seqseq-like neural networks with local and global attention. \newcite{AwI} and \newcite{HierarchicalDialog} design hierarchical neural networks for multi-turn conversation.

To address the problem of universal replies, \newcite{diversity} propose a mutual information training objective. \newcite{variationalDialog} apply a variational Bayes approach that imposes a probabilistic distribution on the hidden variables and encodes the parameters of the posterior distribution.
A very recent study similar to ours is \newcite{topic}, where replies are augmented with topic information. In a dialogue-like question-answering system, \newcite{genQA} query a knowledge base and insert a selected triple into an answer sentence by a soft logistic unit. In such approach, however, the answer may not actually appear in the generated sentence, especially when the test patterns are different from training ones. Unlike existing work, our \seqbf\ model guarantees the predicted keyword can appear in the reply at an appropriate position.

\section{Conclusion and Future Work}\label{sec:conclusion}
In this paper, we proposed a content-introducing approach to generative short-text conversation systems. Instead of generating a reply sequentially from the beginning word to the end as in existing approaches, we used pointwise mutual information to predict a keyword, i.e., a noun term, for the reply.
Then we proposed a ``sequence to backward and forward sequences'' (\seqbf) model to generate a reply containing the predicted keyword. The \seqbf\ mechanism ensures the keyword can appear at an arbitrary position in the reply, but the generated utterance is still fluent. Experimental results show that our approach consistently outperforms the pure \seqseq\ model in dialogue systems in terms of human evaluation and the entropy measure.

In future work, we would like to apply different keyword prediction techniques (e.g., neural sentence models) to improve the performance; the proposed \seqbf\ model can also be extended to other applications like generative question answering, where the answer may be given by searching an external database or knowledge base.

\section*{Acknowledgments}
We thank Jiexiu Zhai, Hao Meng, Siqi Wang, Zhexi Hong, Junling Chen, and Zhiliang Tian for evaluating our dialogue systems.
We also thank all reviewers for their constructive comments. This research is supported by the National Basic Research Program of China (the 973 Program) under Grant No.~2015CB352201 and the National Natural Science Foundation of China under Grant Nos.~61232015, 91318301, 61421091, 61225007, and 61502014.

\bibliographystyle{acl}
\bibliography{seqBF}

\begin{thebibliography}{}

\bibitem[\protect\citename{Bahdanau \bgroup et al.\egroup }2015]{attention}
Dzmitry Bahdanau, Kyunghyun Cho, and Yoshua Bengio.
\newblock 2015.
\newblock Neural machine translation by jointly learning to align and
  translate.
\newblock In {\em Proceedings of the International Conference on Learning
  Representations}.

\bibitem[\protect\citename{Cho \bgroup et al.\egroup }2014]{GRU}
Kyunghyun Cho, Bart van Merri\"enboer, Dzmitry Bahdanau, and Yoshua Bengio.
\newblock 2014.
\newblock On the properties of neural machine translation: {Encoder}-decoder
  approaches.
\newblock In {\em Proceedings of 8th Workshop on Syntax, Semtnatics and
  Structure in Statistical Translation}, pages 103--111.

\bibitem[\protect\citename{Ferguson \bgroup et al.\egroup }1996]{train}
George Ferguson, James Allen, and Brad Miller.
\newblock 1996.
\newblock {TRAINS-95: T}owards a mixed-initiative planning assistant.
\newblock In {\em Proceedings of Artificial Intelligence Planning Systems
  Conference}, pages 70--77.

\bibitem[\protect\citename{Fleiss}1971]{kappa}
Joseph~L Fleiss.
\newblock 1971.
\newblock Measuring nominal scale agreement among many raters.
\newblock {\em Psychological Bulletin}, 76(5):378.

\bibitem[\protect\citename{Graesser \bgroup et al.\egroup }2005]{tutor}
Arthur~C Graesser, Patrick Chipman, Brian~C Haynes, and Andrew Olney.
\newblock 2005.
\newblock {AutoTutor: A}n intelligent tutoring system with mixed-initiative
  dialogue.
\newblock {\em IEEE Transactions on Education}, 48(4):612--618.

\bibitem[\protect\citename{Han \bgroup et al.\egroup }2015]{7patterns}
Sangdo Han, Jeesoo Bang, Seonghan Ryu, and Gary~Geunbae Lee.
\newblock 2015.
\newblock Exploiting knowledge base to generate responses for natural language
  dialog listening agents.
\newblock In {\em Proceedings of the 16th Annual Meeting of the Special
  Interest Group on Discourse and Dialogue}, pages 129--133.

\bibitem[\protect\citename{Isbell \bgroup et al.\egroup }2000]{retrieval1}
Charles~Lee Isbell, Michael Kearns, Dave Kormann, Satinder Singh, and Peter
  Stone.
\newblock 2000.
\newblock {Cobot in LambdaMOO: A} social statistics agent.
\newblock In {\em Proceedings of the 17th National Conference on Artificial
  Intelligence}, pages 36--41.

\bibitem[\protect\citename{Karpathy \bgroup et al.\egroup }2015]{visualizeRNN}
Andrej Karpathy, Justin Johnson, and Fei-Fei Li.
\newblock 2015.
\newblock Visualizing and understanding recurrent networks.
\newblock {\em arXiv preprint arXiv:1506.02078}.

\bibitem[\protect\citename{Li \bgroup et al.\egroup }2016a]{diversity}
Jiwei Li, Michel Galley, Chris Brockett, Jianfeng Gao, and Bill Dolan.
\newblock 2016a.
\newblock A diversity-promoting objective function for neural conversation
  models.
\newblock In {\em Proceedings of the Conference of the North American Chapter
  of the Association for Computational Linguistics: Human Language
  Technologies}, pages 110--119.

\bibitem[\protect\citename{Li \bgroup et al.\egroup }2016b]{stalematebreaker}
Xiang Li, Lili Mou, Rui Yan, and Ming Zhang.
\newblock 2016b.
\newblock {StalemateBreaker: A} proactive content-introducing approach to
  automatic human-computer conversation.
\newblock In {\em Proceedings of the 25th International Joint Conference on
  Artificial Intelligence}, pages 2845--2851.

\bibitem[\protect\citename{Misu and Kawahara}2007]{museum}
Teruhisa Misu and Tatsuya Kawahara.
\newblock 2007.
\newblock Speech-based interactive information guidance system using
  question-answering technique.
\newblock In {\em Proceedings of the International Conference on Acoustics,
  Speech and Signal Processing}, volume~IV, pages 145--148.

\bibitem[\protect\citename{Mou \bgroup et al.\egroup }2015]{BF}
Lili Mou, Rui Yan, Ge~Li, Lu~Zhang, and Zhi Jin.
\newblock 2015.
\newblock Backward and forward language modeling for constrained natural
  language generation.
\newblock {\em arXiv preprint arXiv:1512.06612}.

\bibitem[\protect\citename{Peng \bgroup et al.\egroup }2015]{comparative}
Hao Peng, Lili Mou, Ge~Li, Yunchuan Chen, Yangyang Lu, and Zhi Jin.
\newblock 2015.
\newblock A comparative study on regularization strategies for embedding-based
  neural networks.
\newblock In {\em Proceedings of the Conference on Empirical Methods in Natural
  Language Processing}, pages 2106--2111.

\bibitem[\protect\citename{Ritter \bgroup et al.\egroup }2011]{SMTdialog}
Alan Ritter, Colin Cherry, and William~B Dolan.
\newblock 2011.
\newblock Data-driven response generation in social media.
\newblock In {\em Proceedings of the Conference on Empirical Methods in Natural
  Language Processing}, pages 583--593.

\bibitem[\protect\citename{Rush \bgroup et al.\egroup
  }2015]{seq2seqSummarization}
Alexander~M. Rush, Sumit Chopra, and Jason Weston.
\newblock 2015.
\newblock A neural attention model for abstractive sentence summarization.
\newblock In {\em Proceedings of the Conference on Empirical Methods in Natural
  Language Processing}, pages 379--389.

\bibitem[\protect\citename{Serban \bgroup et al.\egroup
  }2016a]{HierarchicalDialog}
Iulian~V Serban, Alessandro Sordoni, Yoshua Bengio, Aaron Courville, and Joelle
  Pineau.
\newblock 2016a.
\newblock Building end-to-end dialogue systems using generative hierarchical
  neural network models.
\newblock In {\em Proceedings of the 30th AAAI Conference on Artificial
  Intelligence}, pages 3776--3783.

\bibitem[\protect\citename{Serban \bgroup et al.\egroup
  }2016b]{variationalDialog}
Iulian~Vlad Serban, Alessandro Sordoni, Ryan Lowe, Laurent Charlin, Joelle
  Pineau, Aaron Courville, and Yoshua Bengio.
\newblock 2016b.
\newblock A hierarchical latent variable encoder-decoder model for generating
  dialogues.
\newblock {\em arXiv preprint arXiv:1605.06069}.

\bibitem[\protect\citename{Shang \bgroup et al.\egroup }2015]{NRM}
Lifeng Shang, Zhengdong Lu, and Hang Li.
\newblock 2015.
\newblock Neural responding machine for short-text conversation.
\newblock In {\em Proceedings of the 53rd Annual Meeting of the Association for
  Computational Linguistics and the 7th International Joint Conference on
  Natural Language Processing}, pages 1577--1586.

\bibitem[\protect\citename{Sordoni \bgroup et al.\egroup }2015]{BoWdialog}
Alessandro Sordoni, Michel Galley, Michael Auli, Chris Brockett, Yangfeng Ji,
  Margaret Mitchell, Jian-Yun Nie, Jianfeng Gao, and Bill Dolan.
\newblock 2015.
\newblock A neural network approach to context-sensitive generation of
  conversational responses.
\newblock In {\em Proceedings of the Conference of the North American Chapter
  of the Association for Computational Linguistics: Human Language
  Technologies}, pages 196--205.

\bibitem[\protect\citename{Sutskever \bgroup et al.\egroup }2014]{seq2seq}
Ilya Sutskever, Oriol Vinyals, and Quoc~V Le.
\newblock 2014.
\newblock Sequence to sequence learning with neural networks.
\newblock In {\em Advances in Neural Information Processing Systems}, pages
  3104--3112.

\bibitem[\protect\citename{Vinyals and Le}2015]{NNdialog1}
Oriol Vinyals and Quoc Le.
\newblock 2015.
\newblock A neural conversational model.
\newblock {\em arXiv preprint arXiv:1506.05869}.

\bibitem[\protect\citename{Wang \bgroup et al.\egroup }2013]{retrieval2}
Hao Wang, Zhengdong Lu, Hang Li, and Enhong Chen.
\newblock 2013.
\newblock A dataset for research on short-text conversations.
\newblock In {\em Proceedings of the Conference on Empirical Methods in Natural
  Language Processing}, pages 935--945.

\bibitem[\protect\citename{Wang \bgroup et al.\egroup }2016]{poem}
Zhe Wang, Wei He, Hua Wu, Haiyang Wu, Wei Li, Haifeng Wang, and Enhong Chen.
\newblock 2016.
\newblock Chinese poetry generation with planning based neural network.
\newblock In {\em Proceedings the 26th International Conference on
  Computational Linguistics}.

\bibitem[\protect\citename{Xing \bgroup et al.\egroup }2016]{topic}
Chen Xing, Wei Wu, Yu~Wu, Jie Liu, Yalou Huang, Ming Zhou, and Wei-Ying Ma.
\newblock 2016.
\newblock Topic augmented neural response generation with a joint attention
  mechanism.
\newblock {\em arXiv preprint arXiv:1606.08340}.

\bibitem[\protect\citename{Yao \bgroup et al.\egroup }2015]{AwI}
Kaisheng Yao, Geoffrey Zweig, and Baolin Peng.
\newblock 2015.
\newblock Attention with intention for a neural network conversation model.
\newblock {\em arXiv preprint arXiv:1510.08565}.

\bibitem[\protect\citename{Yin \bgroup et al.\egroup }2016]{genQA}
Jun Yin, Xin Jiang, Zhengdong Lu, Lifeng Shang, Hang Li, and Xiaoming Li.
\newblock 2016.
\newblock Neural generative question answering.
\newblock In {\em Proceedings of the 25th International Joint Conference on
  Artificial Intelligence}, pages 2972--2978.

\end{thebibliography}

\end{CJK*}
\end{document}